\documentclass[conference,letterpaper]{IEEEtran}

\usepackage{amssymb,amsmath}
\usepackage{cite} 
\usepackage{booktabs}
\usepackage[font=footnotesize]{caption}
\usepackage{svg}
\usepackage{balance}

\newcommand{\figref}[1]{Fig.\ref{#1}}

\usepackage{hyperref}
\hypersetup{bookmarksopen,bookmarksnumbered,
pdfpagemode=UseOutlines,
colorlinks=true,
linkcolor=blue,
anchorcolor=blue,
citecolor=blue,
filecolor=blue,
menucolor=blue,
urlcolor=blue
}

\newcommand{\xxnote}[3]{}
\ifx\hidenotes\undefined
  \usepackage{color}
  \renewcommand{\xxnote}[3]{\color{#2}{#1: #3}}
\fi

\usepackage{graphicx}

\IEEEoverridecommandlockouts

\begin{document}
{\title{
MuSHR\thanks{This  work  was  partially  funded  by  the Honda Robotics Institute USA, Intel, and the Paul G. Allen School of Computer Science \& Engineering at the University of Washington. We are grateful for their support.}: A Low-Cost, Open-Source Robotic Racecar for Education and Research
}}
\author{
\IEEEauthorblockN{
Siddhartha S. Srinivasa, Patrick Lancaster, Johan Michalove, Matt Schmittle, Colin Summers,
Matthew Rockett,\\ Rosario Scalise, Joshua R. Smith, Sanjiban Choudhury, Christoforos Mavrogiannis, Fereshteh Sadeghi
}
\IEEEauthorblockA{
Paul G. Allen School of Computer Science \& Engineering\\
University of Washington\\
Seattle, WA, 98195-2355 \\
\texttt{mushr@cs.washington.edu}\\
\url{https://mushr.io}\\
}
}

\maketitle

\begin{abstract}
We present MuSHR, the Multi-agent System for non-Holonomic Racing. MuSHR is a low-cost, open-source robotic racecar platform for education and research, developed by the Personal Robotics Lab in the Paul G. Allen School of Computer Science \& Engineering at the University of Washington. MuSHR aspires to contribute towards democratizing the field of robotics as a low-cost platform that can be built and deployed by following detailed, open documentation and do-it-yourself tutorials. A set of demos and lab assignments developed for the Mobile Robots course at the University of Washington provide guided, hands-on experience with the platform, and milestones for further development. MuSHR is a valuable asset for academic research labs, robotics instructors, and robotics enthusiasts.
\end{abstract}

\section{Introduction}\label{sec:intro}

\begin{figure}
    \centering
    \includegraphics[width = \linewidth]{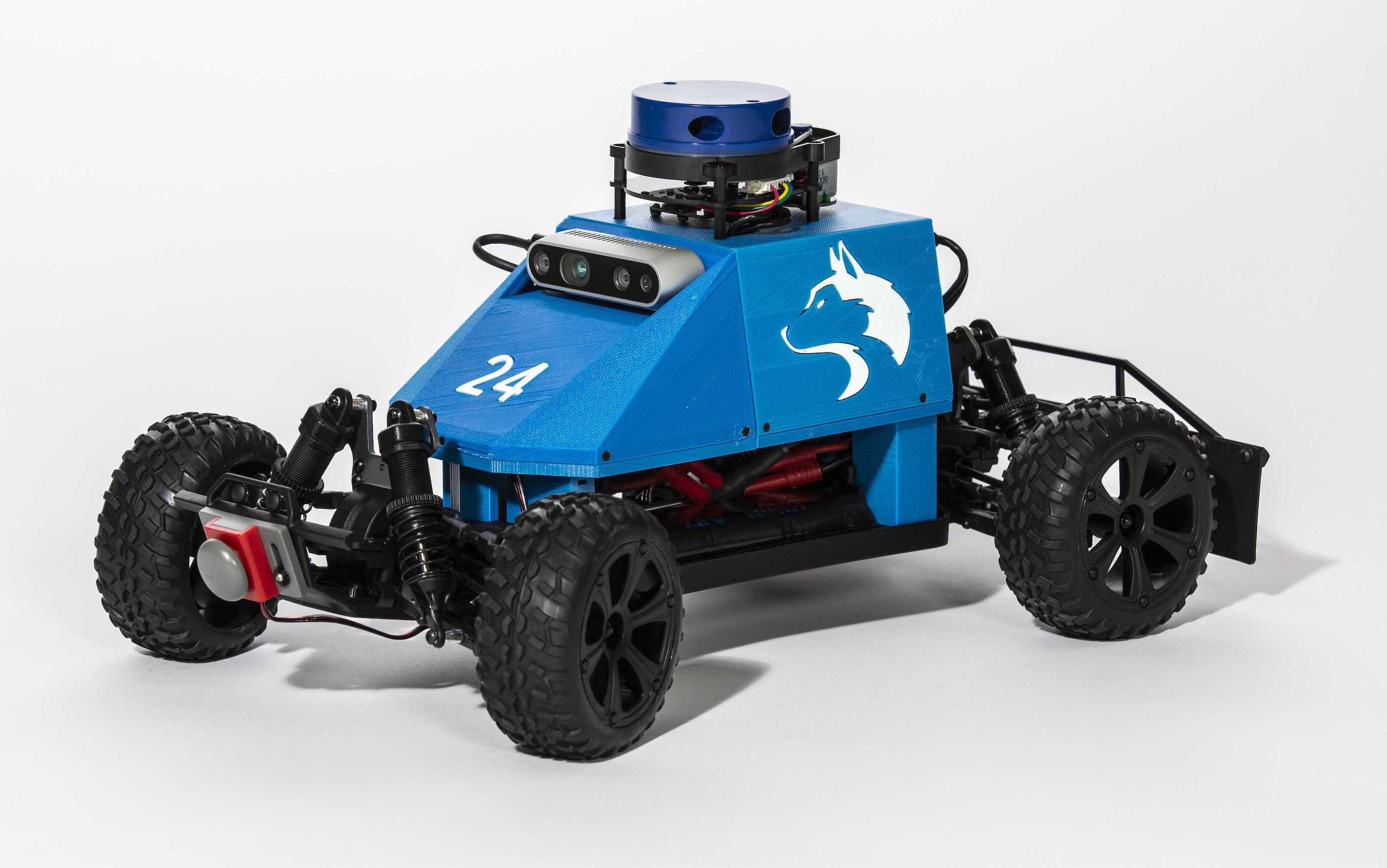}
    \caption{The MuSHR racecar. MuSHR can be built for \$600, through the use of off-the-shelf components, and user-friendly build instructions, found at our website, (\url{https://mushr.io}). A complete open-source software stack can be found at our Github pages, ((\url{https://github.com/prl-mushr})).}
    \label{fig:single-mushr}
\end{figure}

The Multi-agent System for non-Holonomic Racing (MuSHR\footnote{The acronym MuSHR is inspired by dog-sled racing, “mushing”, where dogs (most commonly Alaskan Huskies) work together to pull a sled. As the University of Washington's mascot is the Husky, we found this name especially fitting.}) is an open-source, full-stack robotics platform designed to advance robotics research and education by making a fully integrated robotic race car available in an easy-to-assemble and economic package (\figref{fig:single-mushr} depicts a MuSHR car prototype). Using rapid-prototyping techniques and off-the-shelf parts, we provide an open design that can be built by following do-it-yourself instructions. We also provide an ever-expanding set of tutorials that can guide any user--from the hobby enthusiast to the experienced researcher--through the capabilities of the racecar and underlying robotics principles. The MuSHR platform was developed in the Personal Robotics Lab at the University of Washington's Paul G. Allen School of Computer Science \& Engineering. Taking inspiration from the MIT RACECAR project \cite{mit_racecar}, we set out to create a more affordable (the current design that includes a 2D laser scanner, a RGBD camera, and an IMU can be built for around \$900) full-stack robotics system which not only can support our own research and teaching demands, but those of the community at-large.

\begin{figure*}
    \centering
    \includegraphics[width = \linewidth]{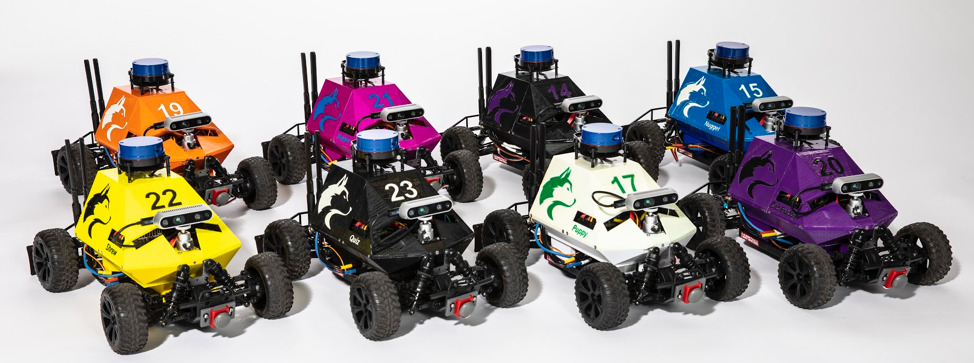}
    \caption{Fleet of MuSHR cars built in the Personal Robotics Lab.}
    \label{fig:mushr_fleet}
\end{figure*}

\section{Platform}

In this section, we provide an overview of the hardware and software architectures of the MuSHR platform. The hardware design is based on a series of off-the-shelf components that can be easily found online and in hardware stores around the world, whereas the software architecture was developed at the Personal Robotics Lab and is provided through our Github page. 

\begin{figure}
    \centering
    \includegraphics[width = \linewidth]{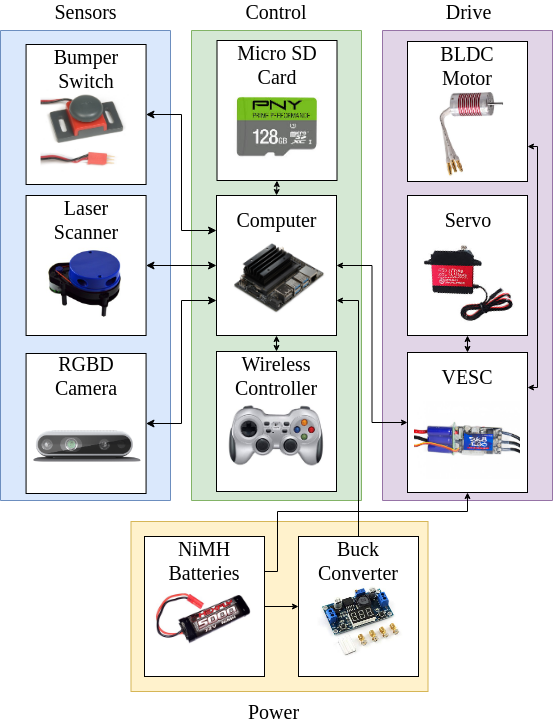}
    \caption{Overview of the MuSHR hardware components.}
    \label{fig:hardware}
\end{figure}

\subsection{Hardware Architecture}

The car is built on a Redcat Racing Blackout SC $1/10$ chassis featuring a 4x4 suspension, and non-flat tires. The chassis accommodates and protects all of the sensing, control, steering, and power subsystems of the car (see \figref{fig:hardware} for an overview of the main system components). The car is equipped with a variety of sensors: an RGBD camera (Intel Realsense D435i), a Laser scanner (YLIDAR X4) providing distance measurements, and a bump sensor (VEX Bumper switch) detecting collisions. Computations take place on a Nvidia Jetson Nano computer, which can easily be loaded with the desired operating system and programs through an SD card. A Logitech F710 wireless controller can also issue commands to the car and may be used for teleoperation. All four wheels are driven by a brushless DC motor (Jrelecs F540 3930KV), whereas a servo motor (ZOSKAY 1X DS3218) controls steering. The whole vehicle is powered by two NiMH batteries (Redcat Racing HX-5000MH-B), one dedicated to powering the sensors and computer, and one used to power the motors. A power converter (DZS Elec LM2596) converts the higher voltage of the battery to the necessary 5V max for the computer, whereas a VESC speed controller (Turnigy SK8-ESC) converts steering and velocity commands into motor/servo commands.

\begin{figure}
    \centering
    \includegraphics[width = \linewidth]{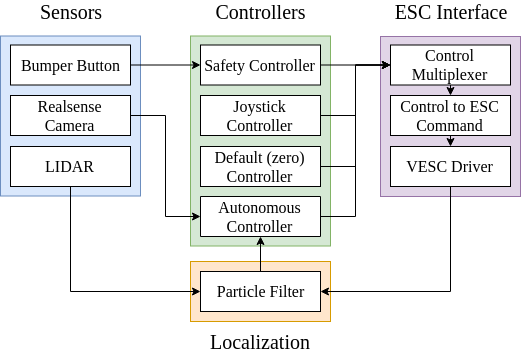}
    \caption{Overview of the MuSHR software architecture.}
    \label{fig:software}
\end{figure}

\subsection{Software Architecture}

The software architecture is depicted in \figref{fig:software}. It comprises four main components: (1) the Sensing interface; (2) the Control module; (3) the Electronic Speed Controller (ESC) interface; (4) the Localization module. We provide a brief description of each one of them:

\paragraph{Sensing interface} Each sensor on the car (see Sensors column of \figref{fig:hardware}) comes with a sensor interface that translates raw inputs into ROS messages. In particular, the \href{https://github.com/EAIBOT/ydlidar}{LIDAR}, and the \href{https://github.com/IntelRealSense/realsense-ros}{Realsense} camera have open source ROS interfaces. The bumper button source takes input from the GPIO pins on the computer and publishes them to a ROS message.
\paragraph{Control Module} An autonomous model predictive controller (\href{https://github.com/prl-mushr/mushr_rhc}{\texttt{mushr\_rhc}}) is provided as an off-the-shelf planner/controller hybrid. \texttt{mushr\_rhc} is flexible to handle both static and dynamic trajectory generation, with a tunable cost function, and \texttt{rviz} debugging visualization tools. Given a map and a goal location, \texttt{mushr\_rhc} plans to nearby waypoints, avoiding mapped obstacles. Additionally, the control module allows for teleoperation via joystick control and has the ability to incoporate a separate a backup safety controller.
\paragraph{Electronic speed controller (ESC) interface} In order to maintain safe operation, the ESC interface multiplexes multiple commands from the teleoperation controller, the autonomous controller, and an optional safety controller. The highest priority commands are sent to the VESC software controller, which smooths them and converts Ackermann steering commands into servo positions and motor speeds.
\paragraph{Localization module} The localization module is based on a particle filter \cite{probabilisticrobotics}, adapted from \cite{mit_racecar}.

The outlined architecture is implemented in our Github navigation stack. Our stack provides a quick, out-of-the-box deployment of a basic functionality, but also makes it easy for users to incorporate additional perception and planning components.

\section{Documentation}

\begin{figure}
    \centering
    \includegraphics[width = \linewidth]{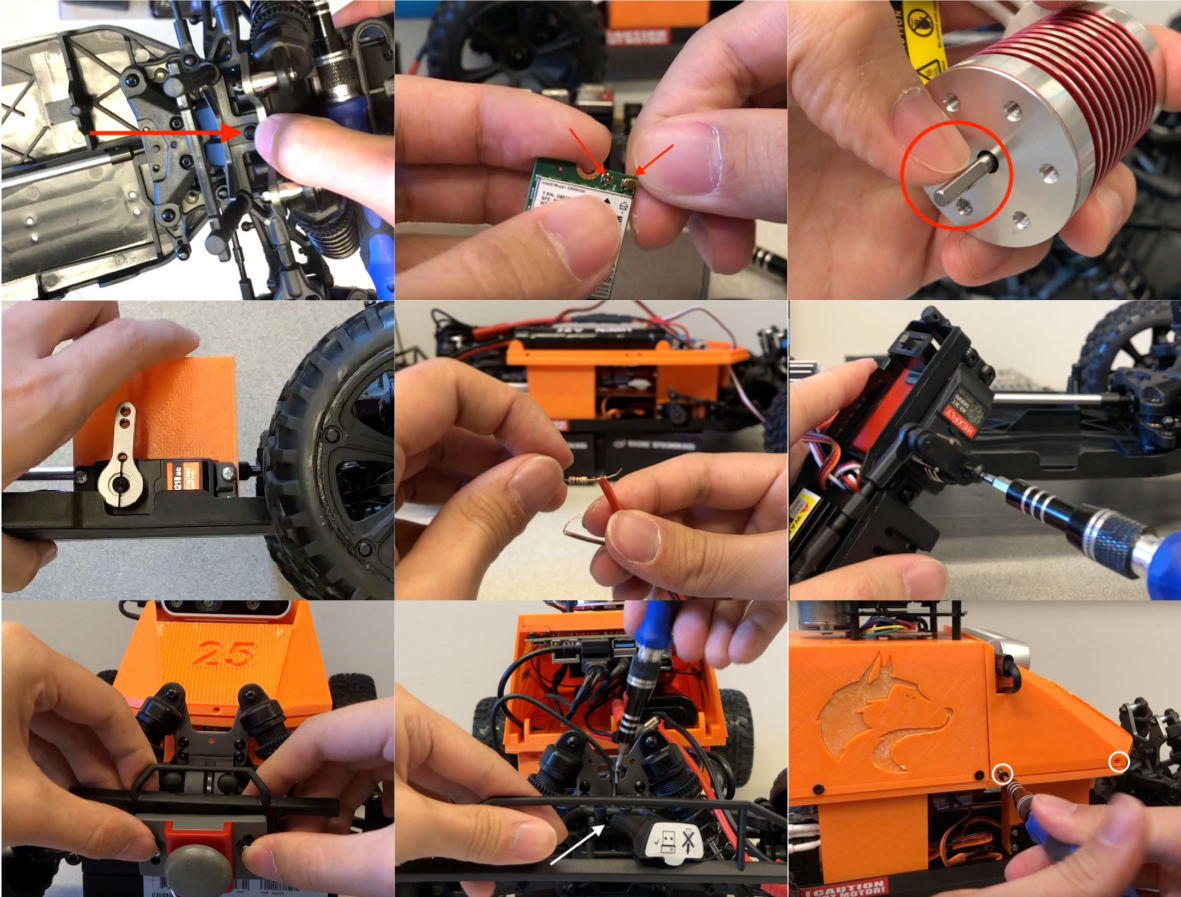}
    \caption{Screenshots from the MuSHR Build Guide, which is available in video format at the MuSHR \href{https://mushr.io/hardware/build_instructions/}{website}.}
    \label{fig:buildinstructions}
\end{figure}

Through the MuSHR webpage (\url{https://mushr.io}) and the MuSHR Github pages (\url{https://github.com/prl-mushr}), we provide open, detailed documentation, including complete building instructions in video format (see \figref{fig:buildinstructions}), manuals for software and hardware components, tutorials, lab assignments from courses at the University of Washington, frequently-asked-questions, and support.

We provide a series of tutorials, guiding the user from their first steps with the platform to more advanced projects, providing time estimates for each milestone. A system overview provides a holistic overview of the racecar software and hardware components, whereas a quick start tutorial enables users to get their platform up and running in as little as 30 minutes. A quick introduction to ROS (the Robot Operating System) \cite{ROS09} familiarizes the user with the fundamentals of modern robot software, whereas more advanced tutorials carry the user through the development of teleoperation and autonomous navigation modules. Finally, a workflow reference provides a set of good practices for building custom components on top of the provided modules and highlights some hints for troubleshooting. We will be enriching and expanding the basic set of tutorials to support the needs of users as the project moves forward.

\section{MuSHR for Education \& Research}

MuSHR is currently the main platform used in \href{https://courses.cs.washington.edu/courses/cse490r/19sp/}{CSE 490R: Mobile Robots}, \href{https://courses.cs.washington.edu/courses/cse571/19wi/}{CSE 571: Algorithms and Applications}, and EE P 545: The Self-Driving Car--Introduction to AI for Mobile Robots at the University of Washington. These courses contain extensive experimental projects on MuSHR, carrying senior undergraduate and graduate students through a series of essential localization, control, and planning algorithms. \figref{fig:mushr_fleet} depicts part of the MuSHR fleet used in the labs of the courses at the University of Washington. We believe that MuSHR could be an invaluable asset for education. From high-school robotics projects, to University-level courses, the low development cost of MuSHR, its detailed documentation and our support, we hope instructors across the globe will benefit from this resource.

The MuSHR platform provides an excellent testbed for showcasing a wide variety of robotics research projects. As examples, we are currently working on a series of exciting research directions including decentralized, multi-robot navigation and collaborative multi-robot manipulation, through the fabrication and attachment of a custom gripper. We are planning to deploy the cars in an indoor workspace, equipped with a high-accuracy, motion-capture system that will allow us to perform robust localization and experiment with a series of interesting, custom-built maps.

\section{Affordability \& Performance}

MuSHR is inspired by the MIT RACECAR \cite{mit_racecar} project but can be built with a fraction of the cost, while not sacrificing autonomous navigation performance. For reference, the basic MuSHR platform, without any sensors can be built with \$600 (a similar MIT racecar setup costs about \$1,000), while a version with a laser scanner and a RGBD camera costs around \$900 (a similar MIT racecar setup costs about \$2,800).

The outlined reduction in the development costs was achieved by incorporating hardware subsystems of lower cost, while ensuring baseline functionalities that could support a wide variety of users, ranging from hobbyists to educators and academic researchers. For example, MuSHR's chassis, including more powerful servo and brushless motors than the MIT car, costs half the price, while not compromising its robustness or controllability. Furthermore, MuSHR makes use of a Jetson Nano processing unit, which costs about three times less than the Jetson TX2 used by the MIT racecar. In practice, we have not found this reduction in processing power to be prohibitive; the robot is still capable of running algorithms for localization, planning, and machine learning. Besides, the use of the Jetson Nano allows for the use of a less complex and expensive power subsystem. In particular, MuSHR’s power sub-system consists of a simple battery and a buck converter, compatible with a wide variety of standard RC car batteries. Finally, MuSHR further reduces costs by using less expensive sensors, such as the YDLIDAR X4 which costs about sixteen times less than the MIT racecar's Hokuyo UST-10LX. We have found that the YDLIDAR X4’s capabilities to be sufficient for typical localization tasks.

\section{Discussion}

This document introduces MuSHR, a low-cost, open-source robotic racecar platform developed by researchers at the Paul G. Allen School of Computer Science \& Engineering at the University of Washington. MuSHR was designed to reach a wide audience, ranging from hobbyists to educators and academic researchers. The platform comes with open-source instructions and video-based tutorials, designed to carry the user through its hardware development. A complete hardware and software documentation introduces the user to the platform, whereas a user guide helps the user get started with the car's basic functionality and covers basic troubleshooting topics. Our entire documentation is hosted on Github, free for everyone to download, use, fork and iterate upon.

\balance
\bibliographystyle{plain}  
\bibliography{references}

\end{document}